\documentclass[10pt, conference, compsocconf]{IEEEtran}
\ifCLASSINFOpdf
\else
\fi
\usepackage{cite}
\usepackage{amsmath,amssymb,amsfonts}
\usepackage{algorithmic}
\usepackage{graphicx}
\usepackage{textcomp}
\usepackage{xcolor}
\def\BibTeX{{\rm B\kern-.05em{\sc i\kern-.025em b}\kern-.08em
		T\kern-.1667em\lower.7ex\hbox{E}\kern-.125emX}}

\usepackage[linesnumbered,ruled,vlined]{algorithm2e}
\usepackage{subcaption}
\usepackage{multirow}

\usepackage[T1]{fontenc} 

\begin{document}
%
\title{Robust Collaborative Learning with Noisy Labels}




%
\author{\IEEEauthorblockN{Mengying Sun\IEEEauthorrefmark{1},
Jing Xing\IEEEauthorrefmark{2},
Bin Chen\IEEEauthorrefmark{2}, 
Jiayu Zhou\IEEEauthorrefmark{1}}
\IEEEauthorblockA{\IEEEauthorrefmark{1}{\it Computer Science and Engineering, Michigan State University, East Lansing, MI, USA}}
\IEEEauthorblockA{\IEEEauthorrefmark{2}{\it Pediatrics and Human Development, Pharmacology and Toxicology, Michigan State University, Grand Rapids, MI, USA}}
{\it Emails: sunmeng2@msu.edu, xingjin1@msu.edu, chenbi12@msu.edu, jiayuz@msu.edu}
}


\maketitle

\begin{abstract}
Learning with curriculum has shown great effectiveness in tasks where the data contains noisy (corrupted) labels, since the curriculum can be used to re-weight or filter out noisy samples via proper design. However, obtaining curriculum from a learner itself without additional supervision or feedback deteriorates the effectiveness due to sample selection bias. Therefore, methods that involve two or more networks have been recently proposed to mitigate such bias. Nevertheless, these studies utilize the collaboration between networks in a way that either emphasizes the disagreement or focuses on the agreement while ignores the other. In this paper, we study the underlying mechanism of how disagreement and agreement between networks can help reduce the noise in gradients and develop a novel framework called Robust Collaborative Learning (RCL) that leverages both disagreement and agreement among networks. We demonstrate the effectiveness of RCL on both synthetic benchmark image data and real-world large-scale bioinformatics data.
\end{abstract}

\begin{IEEEkeywords}
curriculum learning; robust learning; weak supervision

\end{IEEEkeywords}

%
\IEEEpeerreviewmaketitle

\section{Introduction}\label{sec:intro}

Recent years have witnessed huge successes of supervised learning using deep neural networks in various domains \cite{simonyan2014very, devlin2018bert}. One decisive factor behind such successes is the availability of a sufficiently large amount of training data \cite{sun2017revisiting}. In cases where obtaining accurate labels is too expensive, practitioners could use affordable apparatuses to collect less reliable \emph{noisy} labels, e.g., the online crowd-sourcing tool Mechanical Turk. Besides, labels from high-throughput experiments (e.g., biological profiling and chemical screening) often contain inevitable noise due to technical and biological variations. 

Learning with noisy labels has imposed additional challenges. Sometimes the data quality is known \emph{a priori}~\cite{liang2015towards, severyn2015unitn, dehghani2017learning}, but a more common scenario is that, the data available is a mixture of samples with both clean and noisy labels and one does not know, or only has partial knowledge of the underlying distribution of the noise \cite{natarajan2013learning, lin2017focal, menon2015learning, varma2017socratic}. In this problem setting, a learning process that is aware of noise in the labels and actively mitigates the negative impacts from the noisy labels, is the key to improving the generalization of learned models.

Learning with curriculums, eg., self-paced learning (SPL) \cite{kumar2010self, supancic2013self, jiang2014self} reveals its power in dealing with noisy data. The reason is that noisy samples can be re-weighted or even filtered out via proper curriculum designs. It has been proved by \cite{meng2015objective} that the latent objective of self-paced learning is equivalent to a robust loss function, which also shed lights on the effectiveness of SPL on noisy data. However, a major drawback of determining curriculum based on the learner's own ability is the sample selection bias from the learner itself. The error made in early stage will be enhanced as training proceeds. Therefore, two or more networks have been introduced in recent works to mitigate the selection bias \cite{han2018co, lee2019robust}. Nevertheless, these studies either emphasize the disagreement or focus on the agreement between networks without considering the other. Therefore in this paper, we propose a novel framework called \textbf{R}obust \textbf{C}ollaborative \textbf{L}earning (\textbf{RCL}) to deal with noisy labels. The main contributions of this paper are:

\begin{itemize}
	\item We show \textit{disagreement} between networks can diversify the gradients of model weights from noisy samples, which slows down the accumulation of noisy gradients.
	\item We show that under certain conditions, \textit{agreement} from more than one network can improve the quality of data selection, i.e., the label purity increases.
	\item Combining the above two findings, we propose RCL framework that consists of multiple networks, where each network is an individual learner and exchanges knowledge with its \textit{Peer} system. The knowledge of the \textit{Peer} system is fused from multiple networks, by adaptively encouraging \textit{disagreement} in the early stage and \textit{agreement} in the later stage, which fully boost the selection of clean samples for training. 
\end{itemize}

We demonstrate RCL on both synthetic and real experiments. For synthetic experiment, we use the benchmark data CIFAR10 and CIFAR100 \cite{krizhevsky2009learning} under different noise settings following literature \cite{han2018co, van2015learning}. We further validated our framework on cancer drug development using large-scale genomic data \cite{subramanian2017next, woo2019deepcop}. The proposed method achieves state-of-art performance and significantly outperforms baselines in large noise settings, on both image and bioinformatics data.


\section{Related Work}\label{sec:related_work}

Our work originates from curriculum learning and its following variants, and also connects to weak supervision and robust learning. Below are the related works for each direction.


Inspired by the fact that humans learn better when trained with a curriculum-like strategy, \cite{bengio2009curriculum} first proposed curriculum learning. Results on both visual and language tasks have shown that training on easy tasks first and then hard tasks led to faster convergence as well as better generalization. Instead of using a specified curriculum, \cite{kumar2010self} incorporated a latent variable associated with each sample, and jointly optimized the model parameters and the curriculum. Later, a variety of approaches with different predefined curriculums were proposed and validated~\cite{supancic2013self, lee2011learning, jiang2014self, jiang2015self, zhang2016co}. Besides, instead of using predefined curriculum, \cite{jiang2017mentornet} proposed to learn a data-driven curriculum when auxiliary data is available. The authors also designed an efficient algorithm for training very deep neural networks with curriculum. Following that, \cite{han2018co} proposed co-teaching framework, a system of two networks that exchange selected samples to alleviate bias brought by one network. Co-teaching \cite{han2018co} works well empirically with several follow-up works \cite{yu2019does, wang2019co}. \cite{song2019selfie} later proposed to make use of the unselected samples by correcting their labels and combining them with selected samples for training. Other very recent works also aggregated knowledge from multiple sources, e.g., multiple networks or multiple training epochs of a single network to filter out noisy data \cite{lee2019robust, nguyen2019self}.

Learning with corrupted labels also relates to weak supervised learning, and its recent advances can be summarized into the following groups. A common way to leverage weak labels when the quality of data is known \emph{a priori} is to use the pre-train and fine-tune scheme based on the amount of clean and weak data~\cite{deriu2017leveraging, severyn2015unitn, liang2015towards}. Another line of methods design surrogate loss functions for robust learning ~\cite{masnadi2009design, lin2017focal,natarajan2013learning}. Some approaches model the noise pattern or estimate the error transition matrix~\cite{sukhbaatar2014training, menon2015learning}, and denoise by either adding an extra layer~\cite{goldberger2016training}, or using generative models~\cite{ratner2016data, varma2017socratic}. Other methods utilize semi-supervised learning techniques~\cite{abney2004understanding, culp2008iterative} to revise weak labels for further training \cite{han2019deep}, or regularize the learning procedure~\cite{vahdat2017toward}. Recently, learning-to-learn methods have also been proposed to tackle such problems by manipulating gradient update rules~\cite{andrychowicz2016learning, dehghani2017learning}. Since our work mainly follows curriculum learning, we do not expose further details here and refer readers of interest to the original papers.

\section{Background and Investigation}

\subsection{A Revisit of Self-paced Learning (SPL)} 
SPL \cite{kumar2010self} introduces a latent variable associated with each training sample, and solves them during training. Denoting the latent variables in a vector $\mathbf{v} \in [0, 1]^n$, where $n$ is the sample size, the objective function of SPL can be written as:
\begin{align*}
\min_{\mathbf{w, v}\in [0, 1]^n}\mathbf{E}(\mathbf{w, v}, \lambda)=\sum\nolimits_{i=1}^{n}v_iL\big(y_i, f(\mathbf{x}_i, \mathbf{w})\big)+g(v_i, \lambda) 
\end{align*}
where $g(v, \lambda)$ is the curriculum function and regularizes the weight of a given sample, $\lambda$ is a control parameter. $\mathbf{w}$ and $\mathbf{v}$ are optimized alternatively \cite{vaida2005parameter}. A simple example of $g$ can be $g(v, \lambda)=-\lambda v$ with closed-form solution for $v$ at each step:
$v^*(\lambda; l)=1$ when $l < \lambda$ and $v^*(\lambda; l)=0$ otherwise,
where $l$ is the loss for one sample. The design of $g(v, \lambda)$ reveals the nature of SPL: if a sample has large loss on the current model, it is likely to be more difficult to learn or even an outlier.


\subsection{The power of Disagreement}
A major drawback of SPL is the sample selection bias from one learner. The error that takes place in the early stage will be reinforced as training continues. To mitigate this, a second network is introduced in co-teaching \cite{han2018co}, where two networks exchange selected samples to train. Such strategy works better than SPL in practice with the underlying mechanism not well studied. Here, we show that exchanging data introduces disagreement between two networks, which can diversify noisy gradients and lead to higher gradient purity. 


In SGD, the gradient update for each mini-batch is:
\begin{align*}
\mathbf{w}^{t+1} &= \mathbf{w}^t-\eta^t\frac{1}{n_r}\sum\nolimits_{j=1}^{n_r}\nabla l_j(\mathbf{w}^t) = \mathbf{w}^t-\eta^t\frac{1}{n_r}S_{\nabla},
\end{align*}
where $n_r$ is the batch size, $\eta^t$ the step size. Considering each of the two networks, with an oracle that provides the ground truth whether a label is noisy or not, we can decompose gradient summation $S_{\nabla}$ into four disjoint components based on data quality (clean or noisy) and network agreement (agree or not):
\begin{align}
S_{\nabla} = & \underbrace{\Sigma_{j\in I_{11}} \nabla l_j(\mathbf{w}^t)}_{\text{agreed clean}} 
             + \underbrace{\Sigma_{j\in I_{10}} \nabla l_j(\mathbf{w}^t)}_{\text{disagreed clean}} \nonumber\\
            &\quad\quad+ \underbrace{\Sigma_{j\in I_{01}} \nabla l_j(\mathbf{w}^t)}_{\text{agreed noisy}} 
             + \underbrace{\Sigma_{j\in I_{00}} \nabla l_j(\mathbf{w}^t)}_{\text{disagreed noisy}},
\label{eqt:grad_decompose}
\end{align} 
In co-teaching, the set $I_{00}$ (noisy, disagree) are diversified by exchanging data points between the two networks. Such disagreement is crucial and can cause several effects. First, the gradient norm of noisy data may diminish; second, the diversification can take effect across time since gradients are eventually summed and applied to network parameters; third, introducing disagreement is equivalent to adding small perturbations on network parameters, which could increase the robustness of the network.

\begin{figure}[tb]
	\centering
	\begin{subfigure}[b]{0.2\textwidth}
		\includegraphics[width=\textwidth]{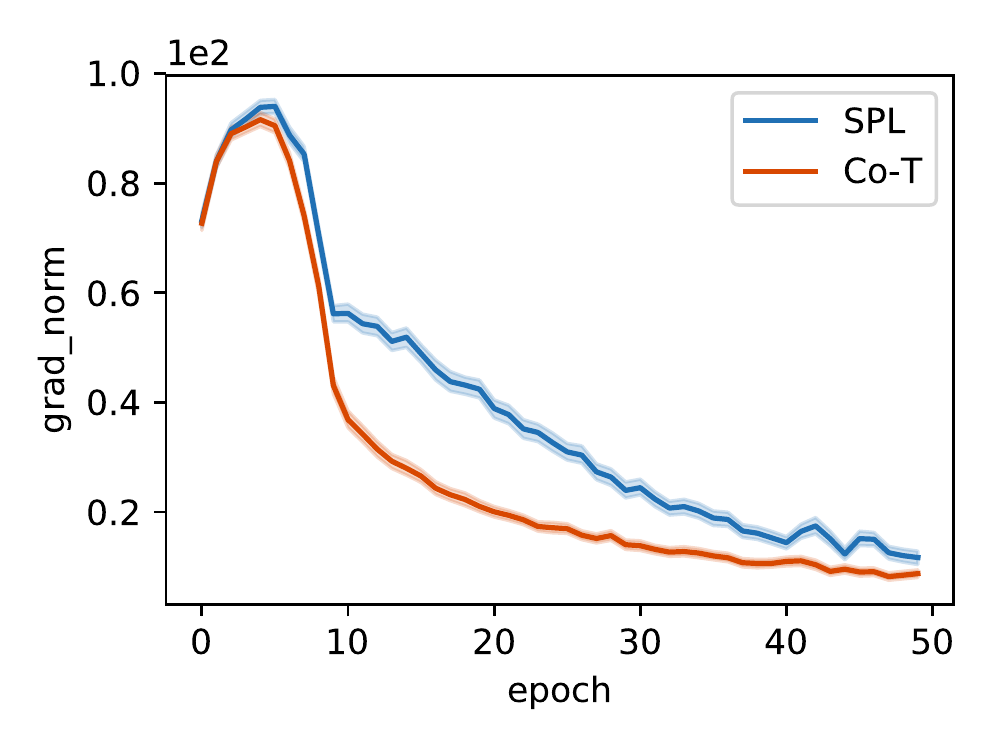}
		\caption{GN at each epoch} \label{gn-1}
		\vspace{-0.05in}
	\end{subfigure}
	~ 
	\begin{subfigure}[b]{0.2\textwidth}
		\includegraphics[width=\textwidth]{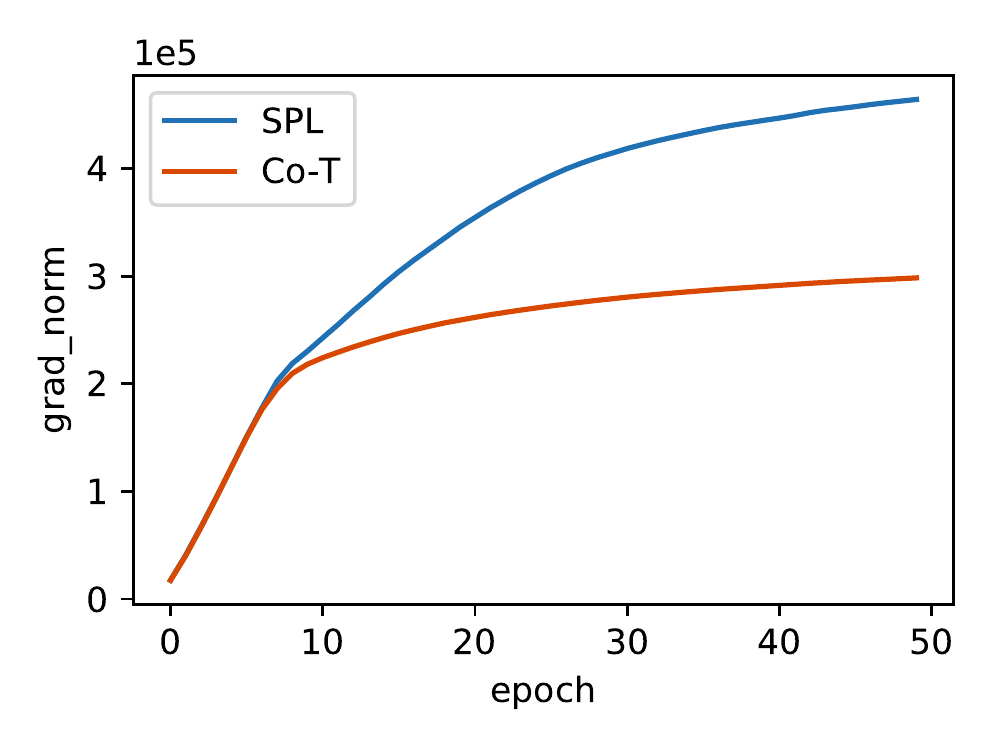}
		\caption{Accumulative GN}\label{gn-2}
		\vspace{-0.05in}
	\end{subfigure}
	\caption{Gradient norm (GN) of \textit{noisy} data on CIFAR10. 
		Co-teaching learning process has less impact from noisy data.
	}\label{grad-norm}
	\vspace{-1.5em}
\end{figure}

We evaluate these effects in a small synthetic experiment. Given an image-classification problem, e.g., CIFAR10, for each class, we manually flip 45\% labels into the adjacent class. Then we compare the gradients of noisy samples between SPL and co-teaching, i.e., $\Sigma_{j\in I_{01}} \nabla l_j(\mathbf{w}^t) + \Sigma_{j\in I_{00}} \nabla l_j(\mathbf{w}^t)$ in Eq. (\ref{eqt:grad_decompose}). The gradients are calculated from the last linear layer of a CNN model. We can see that disagreement from exchanging data helps achieve smaller noisy gradient compared to SPL and it also slows down the accumulation of noisy gradients.




\subsection{The power of Agreement}


When the learners are mature, aggregating their knowledge can be beneficial as compared to only exchanging them. In fact, recent works \cite{lee2019robust, nguyen2019self} propose to aggregate knowledge from multiple networks to filter out noisy samples during training and show promising results. However, ensemble does not guarantee higher purity especially in the early training stage since errors could also be magnified. Therefore, a common strategy is to train the entire data until certain epochs and then perform the ensemble. Nevertheless, such strategy may not be optimal especially when the noise rate is large.

%

\begin{figure}[tb]
	\centering
	\includegraphics[width=0.45\textwidth]{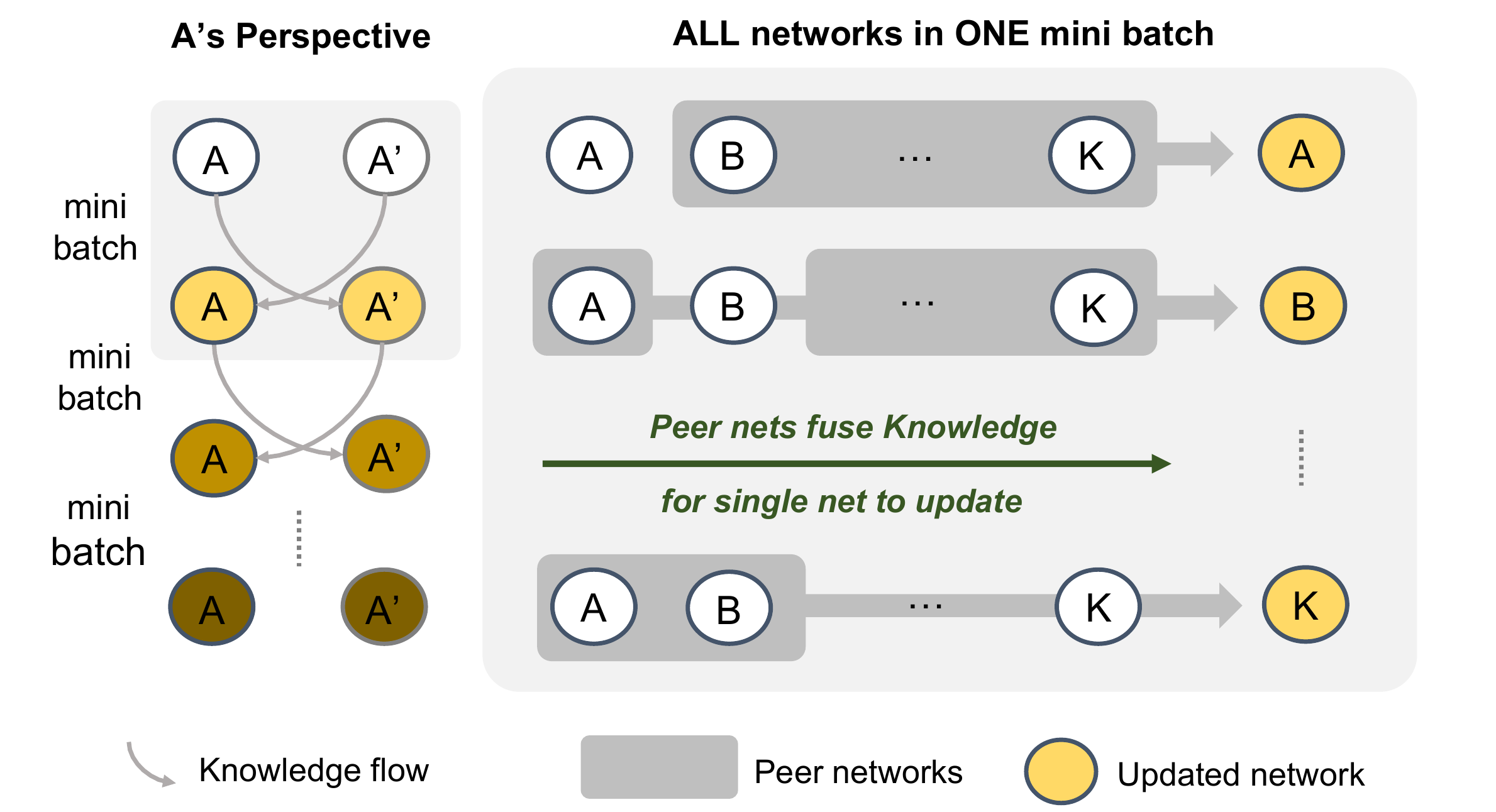}
	\caption{Robust Collaborative Learning (RCL) framework. Left: one network's view, A' denotes all the peer networks of A; Right: all networks in one mini batch. Knowledge fusion and update can be done in parallel for all the networks.}\label{frame-work}
	\vspace{+0.5em}
\end{figure}


\section{Method}\label{sec:method}


Fig.~\ref{frame-work} shows the overall structure of the proposed method. In RCL, each network is an individual learner, while the rest networks form a \textit{Peer} system. From each network's perspective, it exchanges knowledge with its \textit{Peer} (Fig.~\ref{frame-work} Left). The knowledge it receives is fused from multiple networks in the \textit{Peer} system, while the knowledge it offers will wait for fusion when itself is served as a peer network (Fig.~\ref{frame-work} Right). The pseudo code of the overall algorithm is illustrated in Alg.~\ref{alg-1} (real code available at github.com/illidanlab/RCL-code).

\vspace{-0.5em}
\subsection{Self-Knowledge} 
\vspace{-0.5em}
During one mini-batch, each network first selects top $R\times 100\%$ ranked small-loss samples. Due to the memorization effect, i.e., deep neural networks tend to learn easy patterns first and then memorize noise at later epochs~\cite{arpit2017closer}, the reserve rate $R(T)$ is designed to be monotonically decreasing w.r.t. epoch $T$ from $100\%$ until it reaches clean rate $(1-\epsilon) \times 100\%$, where $\epsilon$ is the noise rate. $T_{cut}$ is the switch epoch, after which, only $(1-\epsilon) \times 100\%$ of the data will be selected. The selected samples are the self-knowledge of each individual network, and will be used in knowledge fusion step.

\setlength{\textfloatsep}{0.1pt}
\begin{algorithm}[tb]\small
	\DontPrintSemicolon 
	\KwIn{$K$ networks \{$\Theta_1..\Theta_K$\}, training data $\mathcal{D}$, noise rate $\epsilon$; 
		\\ (Fixed) learning rate $\eta$, epoch $T_{\max}$ and iteration $N_{\max}$;
		\\ (Hyper) epoch $T_{\text{cut}}$, fusion multiplier $\alpha$, fusion exponent $\beta$. }
	\KwOut{Updated network parameters \{$\Theta_1'..\Theta_k'$\}}
	\For{$T = 1$ \textbf{to} $T_{\max}$} {
		\textbf{Shuffle} training set $\mathcal{D}$\\
		\textbf{Update} $R(T)=1-\epsilon \cdot \min\Big\{\frac{T}{T_{\text{cut}}}, 1\Big\}$ \quad  // remember rate\\
		\textbf{Update} $r(T)=1-\min \Big\{\Big(\frac{T}{\alpha T_{\text{cut}}}\Big)^\beta, 1\Big\}$ \quad // fusion rate \\
		\For{$N = 1$ \textbf{to} $N_{\max}$} {
			\textbf{Fetch} mini-batch $D$ from $\mathcal{D}$ \\
			\For{$k = 1$ \textbf{to} $K$}{
				// pick top $R(T)$ small-loss instances\\
				\textbf{Obtain} $D_k=\arg\min_{\mathbb{D}:|\mathbb{D}|\le R(T)|D|}l(f_{\Theta_k},\mathbb{D})$
			}
			\For{$k = 1$ \textbf{to} $K$}{
				// integrate knowledge from all \textbf{peer} networks\\
				\textbf{Obtain} $D_k'=\textit{Knowledge } \Big( D_{\{1..K\} \setminus k}, r(T) \Big)$\\
				// update network $k$\\
				\textbf{Update} $\Theta_k'=\Theta_k-\eta\nabla l(f_{\Theta_k}, D_k')$  	
			}
		}
	}
	\Return{$\Theta' = \{\Theta_1'..\Theta_k'\}$};
	\caption{Pseudo code for RCL Algorithm.}
	\label{alg-1}
\end{algorithm}


\vspace{-0.5em}
\subsection{Knowledge Fusion}
\vspace{-0.5em}
For a given network $k$, it utilizes knowledge from its \textit{Peer} system, which includes all the rest networks except for network $k$. The knowledge of this system is fused from multiple networks via a knowledge fusion function. Ideally, when the networks are trained well, the knowledge of agreement, i.e., data points picked by all the peer networks can be used to update network $k$. However, during the early stage, the networks are prone to making mistakes. Therefore, disagreement is introduced to reduce the noise in gradients. There are two parameters associated with it. $\alpha$ determines the switch epoch between disagreement and agreement. Instead of setting a particular epoch, we design $\alpha$ as the lag of switch epoch compared to $T_{cut}$. Fusion rate $r$ controls the strength of disagreement, i.e., the proportion of disagreed samples that will be included in addition to the common samples. As epoch increases, less disagreed samples will be included until only common ones are selected. The decay of strength of disagreement is controlled by a hyper-parameter $\beta$. In summary, the fusion rate at each epoch is calculated as:
\begin{align}
r(T) = \begin{cases}
1-\left(T/ (\alpha T_{\text{cut}})\right)^\beta, & \text{if } T<\alpha T_{\text{cut}}\\
0, & \text{if } T\ge \alpha T_{\text{cut}}
\end{cases}
\end{align}
After that, $r\times 100\%$ of the disagreed samples are randomly picked and added to the agreed samples as the final knowledge of the \textit{Peer} system. Such randomness also introduces certain level of disagreement since each network will receive different candidates to train even if the common samples are the same within each \textit{Peer} system. The pseudo code of knowledge fusion procedure for multiple networks is illustrated in Alg.~\ref{alg-2}. 

%

\setlength{\textfloatsep}{0.1pt}
\begin{algorithm}[tb]\small
	\DontPrintSemicolon 
	\KwIn{Given the $k$-th network, the knowledge of all other \textbf{peer} networks $\{D_1..D_K\} \setminus D_k$; Fusion rate $r(T)$.}
	\KwOut{Data for updating the $k$-th network $D_k'$.}
	\quad $D_{\text{agree}}=\text{Intersect }\Big(\{D_1..D_K\} \setminus D_k\Big)$\\
	\quad $D_{\text{potential}}=\text{Union }\Big(\{D_1..D_K\} \setminus D_k\Big)$\\
	\If{\quad $|D_{\text{agree}}|==|D_{\text{potential}}|$}{\quad $D_k'=D_{\text{agree}}$}	
	\Else{\quad $D_{\text{uncertain}}=D_{\text{potential}}- D_{\text{agree}}$\\ 
		\quad $n_{\text{in}} = r(T) \cdot |D_{\text{uncertain}}|$\\
		\quad $D_{\text{in}}=\text{random\_sample}\Big(D_{\text{uncertain}}, n_{\text{in}}\Big)$\\
		\quad $D_k'=D_{\text{agree}} + D_{\text{in}}$\\}
	\Return{$D_k'$};
	\caption{Knowledge Fusion Function.}
	\label{alg-2}
\end{algorithm}

\subsection{Knowledge Update}
Network $k$ receives the candidate samples from its \textit{Peer} system and update parameters based on them. The same procedure can be done in parallel for all the networks. After all the networks have been updated, they enter the next iteration.




\begin{table*}[tb]
	\centering
	\begin{tabular}{cccccccccccccc}
		\hline
		Method            & Standard & SPL   & De-CP & Co-T & K=3   & K=5   & K=7   & K=9   & K=11  & K=13  & +R             & p-val    & \#   nets \\ \hline
		Data Noise        & \multicolumn{13}{c}{Test Accuracy}                                                                                                  \\ \hline
		CIFAR10 SYM 50\%  & 48.52    & 70.92 & 45.53      & 73.45       & 75.72 & 77.44 & 78.01 & 78.47 & \textbf{78.91} & 78.91 & \textbf{7.43}  & 2e-9 & 11        \\
		CIFAR10 PF 45\%   & 48.65    & 56.08 & 49.24      & 72.77       & 74.59 & 76.28 & 77.28 & 78.25 & 78.70 & \textbf{79.15} & \textbf{8.77}  & <1e-9    & 13        \\
		CIFAR100 SYM 50\% & 21       & 36.21 & 17.51      & 38.14       & 40.05 & 41.83 & 42.43 & 42.96 & \textbf{43.77} & 43.14 & \textbf{14.75} & 4e-07 & 11        \\
		CIFAR100 PF 45\%  & 29.57    & 28.63 & 26.17      & 30.44       & 32.51 & 34.90 & 36.86 & 37.85 & 39.02 & \textbf{39.15} & \textbf{28.58} & 4e-09 & 13        \\ \hline
		Data Noise        & \multicolumn{13}{c}{Pure Ratio}                                                                                                     \\ \hline
		CIFAR10 SYM 50\%  & 50.31    & 84.22 & 40.48      & 83.95       & 86.96 & 88.82 & 89.47 & 89.86 & 90.11 & \textbf{90.33} & \textbf{7.33}  & <1e-9    & 11        \\
		CIFAR10 PF 45\%   & 54.89    & 68.09 & 51.23      & 79.25       & 82.72 & 84.99 & 86.16 & 87.17 & 87.69 & \textbf{88.11} & \textbf{11.18} & <1e-9    & 13        \\
		CIFAR100 SYM 50\% & 49.95    & 80.51 & 42.89      & 80.65       & 83.85 & 86.20 & 87.14 & 87.82 & \textbf{88.20} & 88.20 & \textbf{9.36}  & <1e-9    & 11        \\
		CIFAR100 PF 45\%  & 54.84    & 58.66 & 53.42      & 59.03       & 61.07 & 63.94 & 66.97 & 68.65 & 69.36 & \textbf{69.99} & \textbf{18.57} & <1e-9    & 13        \\ \hline
	\end{tabular}
	\vspace{-0.5em}
	\caption{Performance of non-ensemble baselines and the proposed method RCL over different number of networks (K) on fixed noise rates. Average over 5 random seeds. +R denotes relative improvement. Significance $t$-tests (one-side) are conducted between RCL and the best baseline. A $p$-value less than 0.05 is considered as significant difference.}\label{num_network}
	\vspace{-1em}
\end{table*}

\section{Experiment}\label{sec:exp}


\subsection{Synthetic Experiment}

\noindent\textbf{Datasets.} We use CIFAR-10 and CIFAR-100 datasets and manually create corrupted labels following the strategy in \cite{han2018co, van2015learning}. Two common noise scenarios are considered, symmetric flip (SYM) and pair flip (PF). For SYM, the label of each class is uniformly random flipped to the rest classes with equal probability; for PF, the label of each class only flips to one different but similar class. The noise rate $\epsilon$ quantifies the overall proportion of labels that are flipped for each class.



\noindent\textbf{Network Architecture.} We follow the 9-layer CNN architecture from \cite{han2018co, laine2016temporal}. We use Adam optimizer with a momentum of 0.9 and an initial learning rate of 0.001. The batch size is 128 and the maximum epoch is 200. We implemented and run the models using PyTorch and NIVIDIA GPUs. 

%

\noindent\textbf{Experimental Setup} We keep the major hyper-parameter of co-teaching~\cite{han2018co}, $T_{\text{cut}}$ in Alg.~\ref{alg-1}, and fix those having subtle effect on results in the original paper. In our experiments, we first tune $T_{\text{cut}}=\{5, 10, 15\}$ for one and two networks (i.e., SPL and Co-teaching). Then we use the best for further tuning RCL. For other hyper-parameters, we fix $\alpha=2$ for all scenarios and use fixed $\beta$ for a given noise scenario, based on the sensitivity analysis using $\beta=\{0.0, 0.1, 0.3, 0.5, 1.0, 2.0, 8.0\}$. We test over a range of number of networks $K=\{3, 5, 7, 9, 11, 13\}$, and noise rates $\epsilon = \{0.25, 0.35, 0.45\}$ for pairflip noise and $\epsilon = \{0.5, 0.6, 0.7, 0.8\}$ for symmetric noise.

\noindent\textbf{Baselines.} (1) Standard, a single network trained on the entire dataset. (2) SPL \cite{jiang2017mentornet}, a single network that produces curriculum based on its own. (3) Decoupling (De-CP) \cite{malach2017decoupling}, a double-net system where the networks only updates parameters from data whose prediction label is disagreed between two networks. (4) Co-teaching \cite{han2018co}, a double-net system in which the two networks exchange curriculum at each iteration. (5) Ensemble consensus \cite{lee2019robust}, specifically the LNEC variant, a multi-net system that explores agreement between multiple networks. (6) Self-ensemble (SELF) \cite{nguyen2019self}, which explores agreement between consecutive epochs within a network. The original implementation involves other hybrid components without code release, we adopt the core idea of temporal ensemble and implement the method. The test accuracy is evaluated on \textit{clean} test set. The pure ratio measures the average proportion of clean data that is selected by the algorithm during training. All metrics are evaluated on \emph{one} network. 



\noindent\textbf{Benefit of agreement.} The benefit of agreement comes from ensemble of multiple networks on the selection of clean samples, which can be verified by adding number of networks while fixing other components. Table \ref{num_network} shows the results of RCL over different number of networks for a given noise rate. We can see that test accuracy improves as the number of networks increases. Higher pure ratio generally leads to higher test accuracy. Importantly, RCL selects significantly more clean data compared to baselines. 

\begin{figure}[tb]
	\centering
	\begin{subfigure}[b]{0.24\textwidth}
		\includegraphics[width=\textwidth]{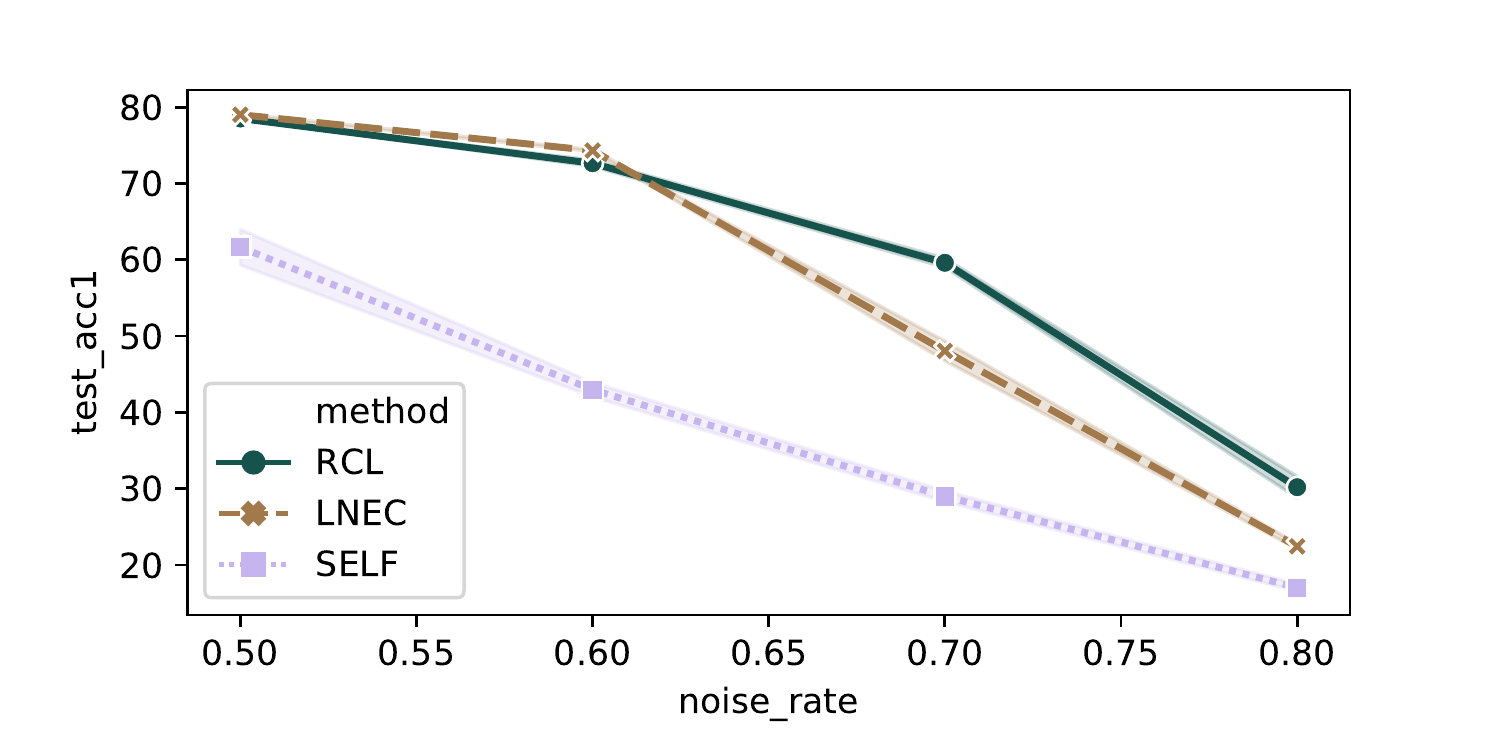}
		\caption{CIFAR10 symmetric flip}
	\end{subfigure}
	\begin{subfigure}[b]{0.24\textwidth}
		\includegraphics[width=\textwidth]{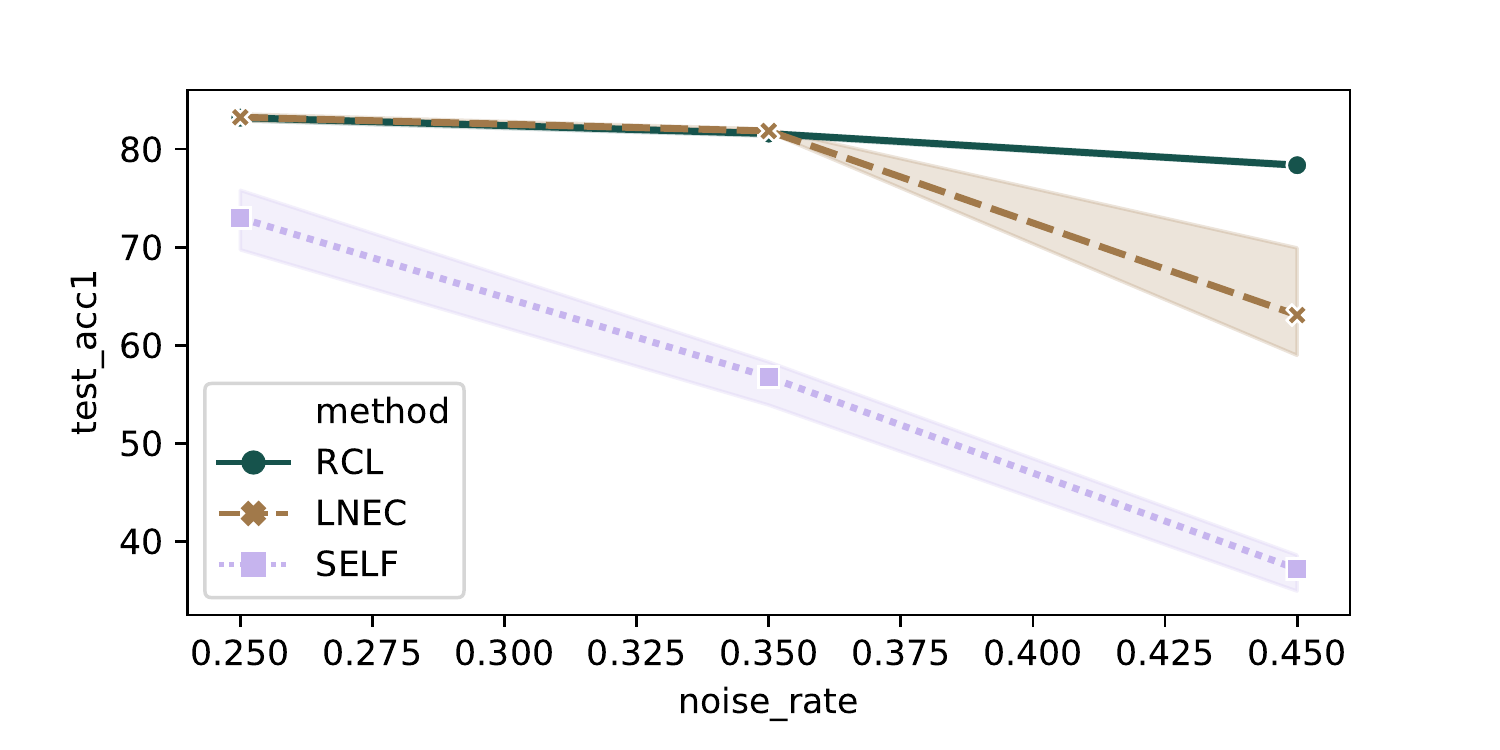}
		\caption{CIFAR10 pair flip}
	\end{subfigure}
	\caption{Test accuracy of ensemble methods and RCL (K=9) over various noise rates on CIFAR10 (avg. 3 runs).}\label{noise-rate}
	\vspace{-0.5em}
\end{figure}

\begin{table}[t]
	\centering
	\begin{tabular}{cccccccc}
		\hline
		\multirow{2}{*}{Data}                                                                & \multirow{2}{*}{Noise} & \multicolumn{3}{c}{Test accuracy} & \multicolumn{3}{c}{Pure ratio} \\ \cline{3-8} 
		&                        & SELF        & LNEC      & RCL        & SELF       & LNEC     & RCL       \\ \hline
		\multirow{4}{*}{\begin{tabular}[c]{@{}c@{}}CF10\\ SYM \end{tabular}} 
		& 50\%                    & 61.65     & 79.00     & 78.51     & 81.38    & 90.07    & 89.85    \\
		& 60\%                    & 42.97     & \textbf{74.32}     & 72.65     & 64.89    & \textbf{86.15}    & 85.14    \\
		& 70\%                    & 28.99     & 48.05     & \textbf{59.60}     & 46.20    & 61.31    & \textbf{74.40}    \\
		& 80\%                    & 17.04     & 22.45     & \textbf{30.20}     & 29.13    & 31.65    & \textbf{42.04}    \\ \hline
		\multirow{3}{*}{\begin{tabular}[c]{@{}c@{}}CF10\\ PF \end{tabular}}          
		& 25\%                   & 72.97     & 83.23     & 83.18     & 88.34    & 93.13    & 92.75    \\
		& 35\%                   & 56.77     & 81.83     & 81.57     & 75.58    & 91.05    & 90.93    \\
		& 45\%                   & 37.21     & 63.09     & \textbf{78.34}     & 57.61    & 72.22    & \textbf{87.28}    \\ \hline
		\multirow{4}{*}{\begin{tabular}[c]{@{}c@{}}CF100\\ SYM \end{tabular}} 
		& 50\%                    & 25.02     & 43.89     & 43.37     & 66.69    & 86.92    & \textbf{87.99}    \\
		& 60\%                    & 15.82     & 32.07     & \textbf{36.12}     & 52.60    & 78.06    & \textbf{82.20}    \\
		& 70\%                    & 7.69      & 23.41     & \textbf{26.71}     & 37.23    & 64.44    & \textbf{72.34}    \\
		& 80\%                    & 3.29      & 11.99     & \textbf{14.79}     & 24.02    & 40.20    &\textbf{49.96}    \\ \hline
		\multirow{3}{*}{\begin{tabular}[c]{@{}c@{}}CF100\\ PF \end{tabular}}          
		& 25\%                   & 38.23     & \textbf{52.83}     & 51.03     & 82.21    & \textbf{92.04}    & 87.94    \\
		& 35\%                   & 28.08     & \textbf{46.49}     & 45.91     & 69.12    & \textbf{82.28}    & 81.28    \\
		& 45\%                   & 19.28     & 31.48     & \textbf{38.38}     & 53.99    & 61.16    & \textbf{69.17}    \\ \hline
	\end{tabular}
	\caption{Final performance of ensemble methods and RCL over various noise rates (K=9). Bold numbers indicate that the method is significantly better than the second best method.} \label{nr-tab}
\end{table}

\begin{figure}[t]
	\vspace{-2em}
	\begin{subfigure}[b]{0.23\textwidth}
		\includegraphics[width=\textwidth]{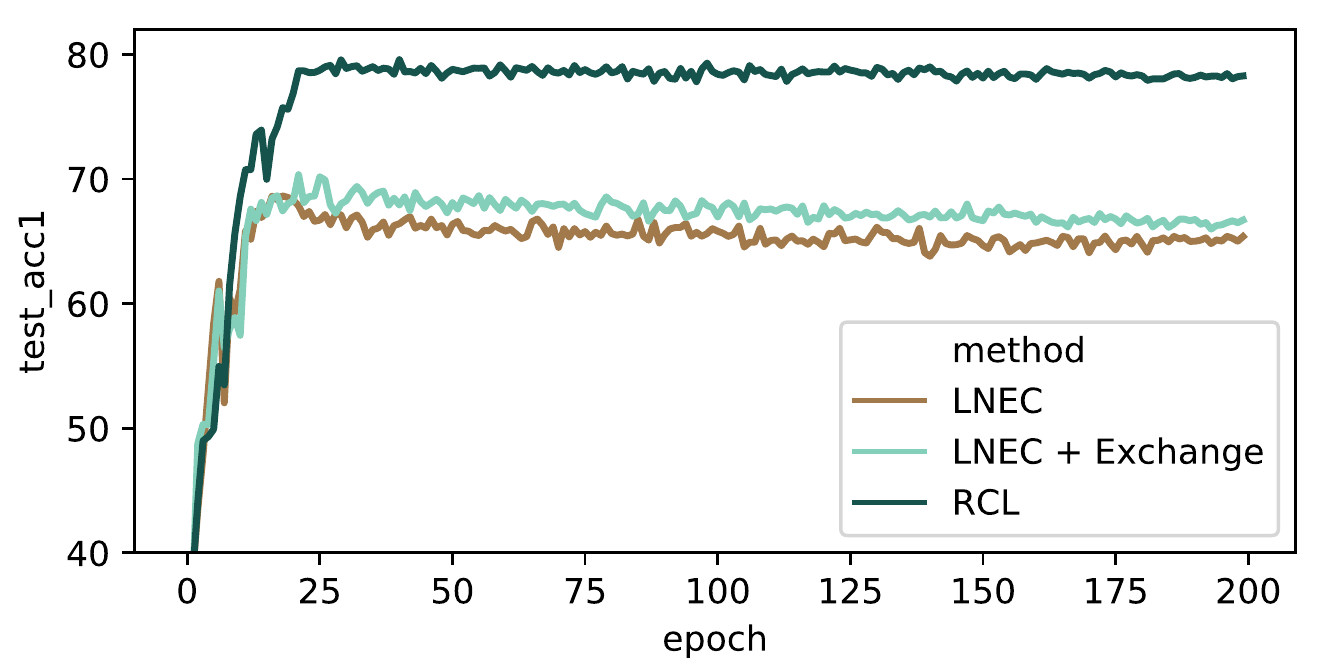}
		\caption{Test accuracy}
	\end{subfigure}
	\begin{subfigure}[b]{0.23\textwidth}
		\includegraphics[width=\textwidth]{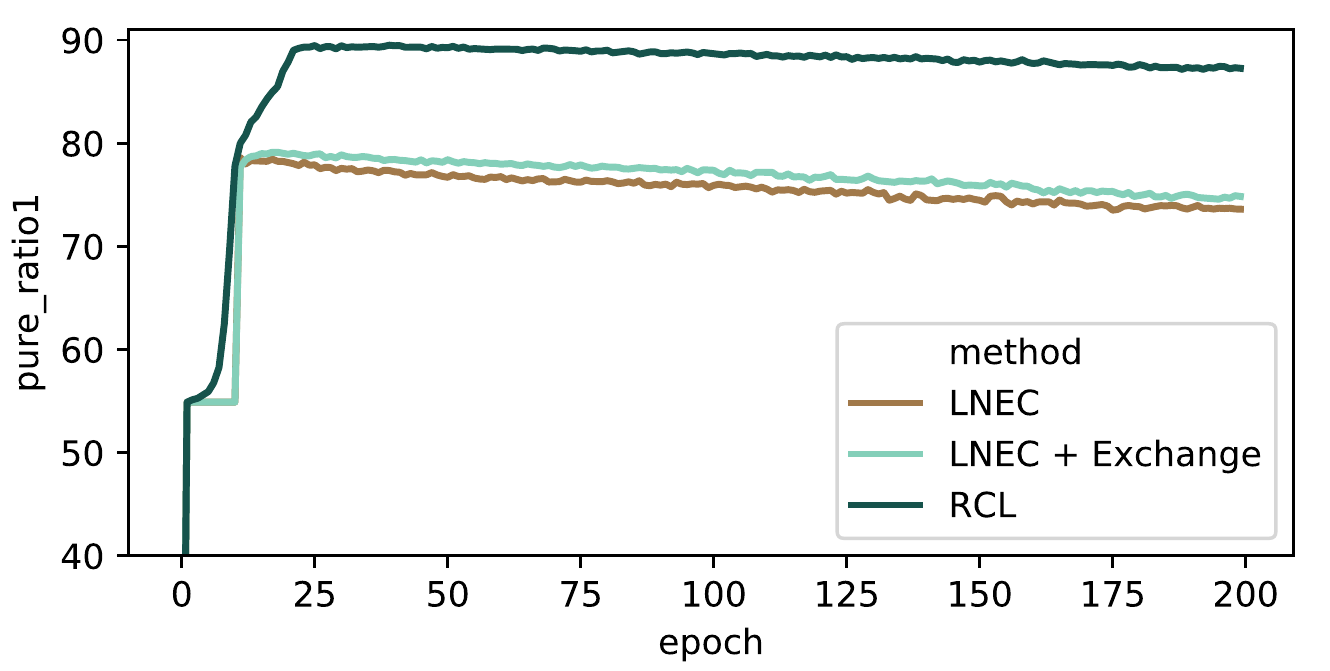}
		\caption{Pure ratio}
	\end{subfigure}
	\caption{Improvement of RCL over ensemble method by each \textit{disagreement} level on CIFAR10 PF 45\% scenario (K=9).}\label{module}
	\vspace{-1em}
\end{figure}

\noindent\textbf{Benefit of disagreement.} The agreement may not work when the noise rate is large, especially during the early training stage, since it also ensembles the error. In such situation, disagreement can help reduces the noise in gradients and helps pick out more clean samples. The disagreement takes place in terms of two levels: the first level is to exchange data in a learn-from-the-other way such that each network receives different data from the its \textit{Peer} system, the second level is that the strength of disagreement varies along the training procedure and can be controlled by a hyper-parameter. Both levels can improve the selection of clean samples as well as the generalization performance. To verify these, we compare RCL with several ensemble methods \cite{lee2019robust, nguyen2019self} which only explore agreement among multiple networks (or multiple training epochs), when all the networks use the same set of candidates to train. Fig. \ref{noise-rate} shows the test accuracy of the competing methods for different noise rates on CIFAR10. Pure ratio reveals the exact same pattern and therefore is not shown in the figure. Complete results are presented in Table \ref{nr-tab}. First, temporal ensemble (SELF) does not work as good as network ensembles in the provided scenarios. Second, when the noise rate is small, RCL reaches similar accuracy as the state-of-art ensemble methods; when the noise rate is large, RCL yields significantly better performance compared to the ensemble baselines. We also find that the variations of baselines are much larger compared to RCL, which indicates the robustness of RCL. Next, in order to see the improvement brought by each level of disagreement, we add the data exchange step onto LNEC baseline and compare it with pure LNEC and RCL (all use 9 networks). The result is shown in Fig. \ref{module}. 

\begin{figure}[tb]
	\centering
	%
	\begin{subfigure}[tb]{0.23\textwidth}
		\includegraphics[width=\textwidth]{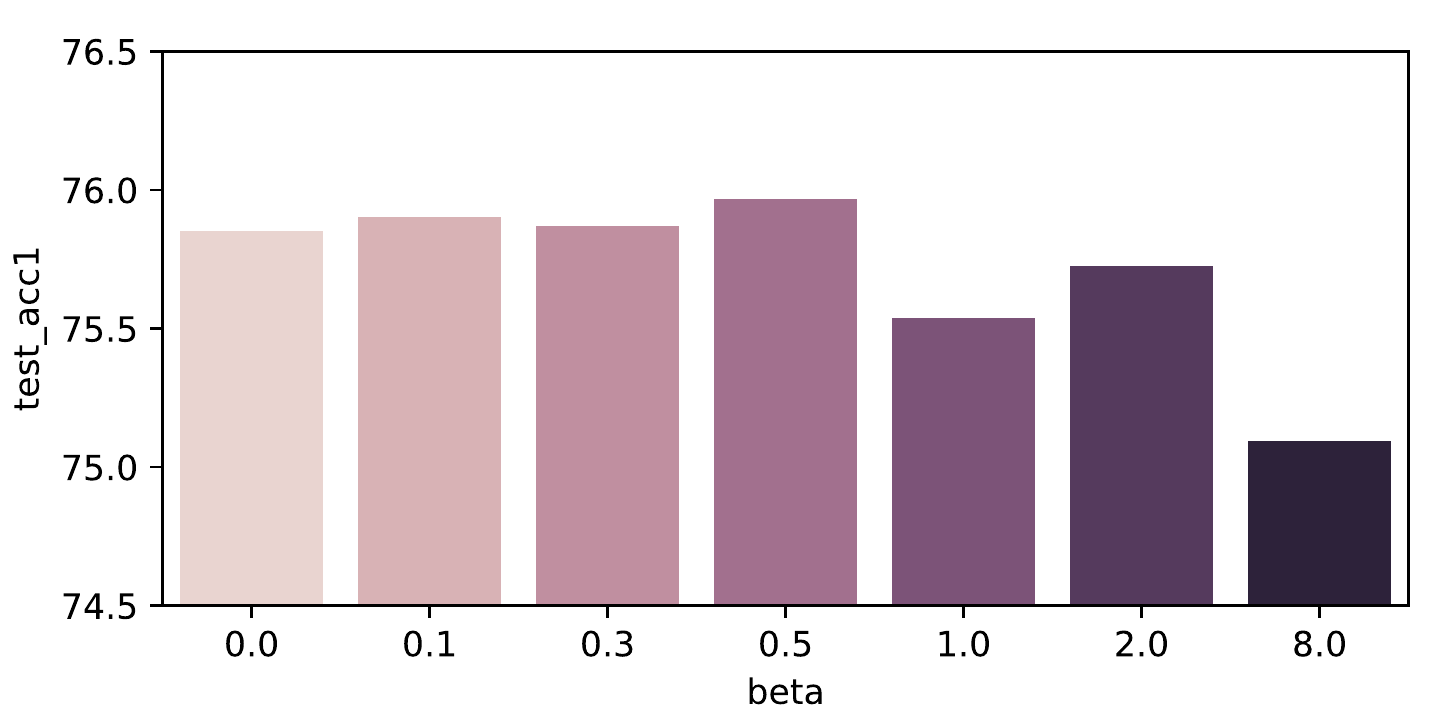}
		\caption{symmetric 50\%}
	\end{subfigure}
	\begin{subfigure}[tb]{0.23\textwidth}
		\includegraphics[width=\textwidth]{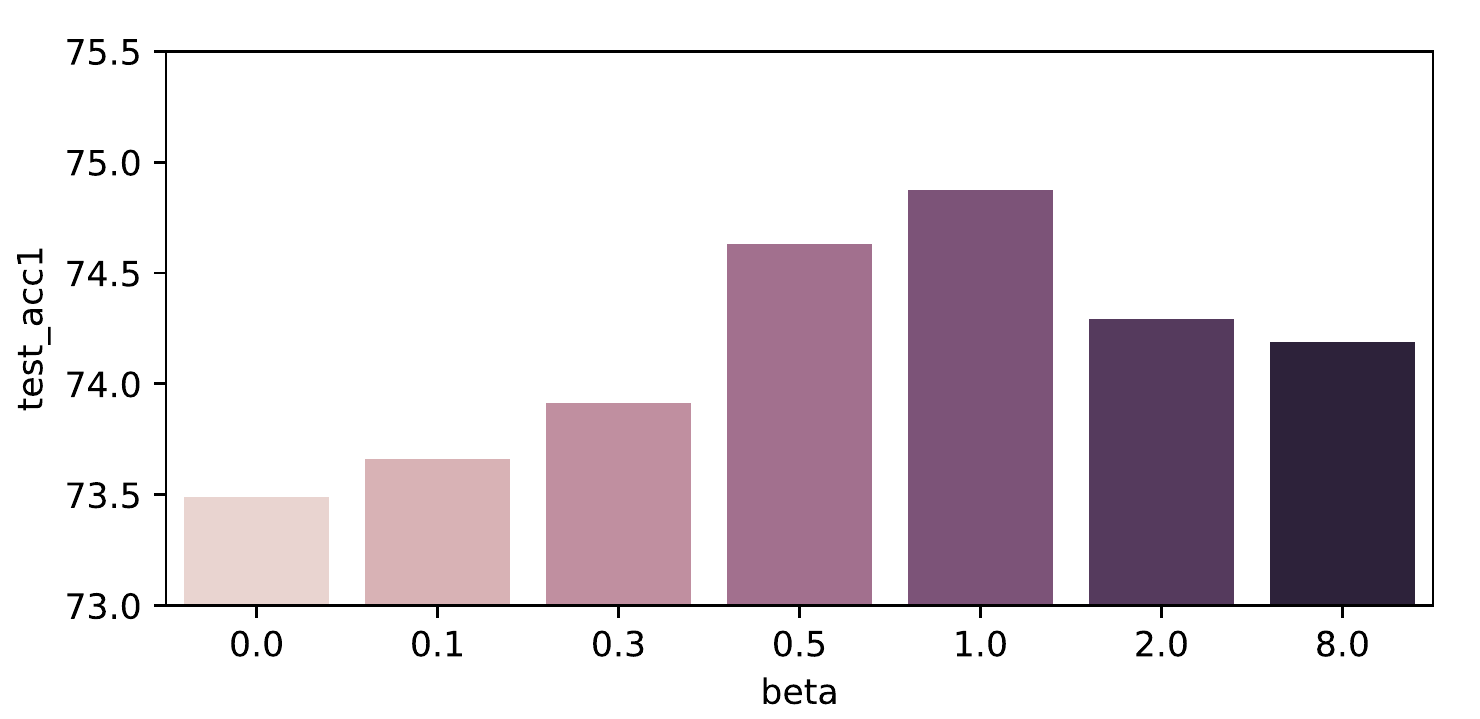}
		\caption{pairflip 45\%}
	\end{subfigure}
	\caption{Test accuracy of RCL on various $\beta$s (avg. 3 runs, K=3).}\label{sensitivity}
\end{figure}


\noindent\textbf{Sensitivity analysis} The parameter $\beta$ controls the strength of disagreement and is important in RCL. We test over a range of different values to study its behavior. Fig. \ref{sensitivity} shows the results on CIFAR10. We can see the test accuracy shows opposite patterns for different noise scenarios. When the task is relatively easy (e.g., CIFAR10 symmetric 50\%), small $\beta$ yields better accuracy. If the noise rate is large, it favors relatively large $\beta$ which encourages disagreement during the early stage. However, extreme large value of $\beta$ is not beneficial. 

\begin{figure}[tb]
	\vspace{-2em}
	\centering
	\begin{subfigure}[b]{0.23\textwidth}
		\includegraphics[width=\textwidth]{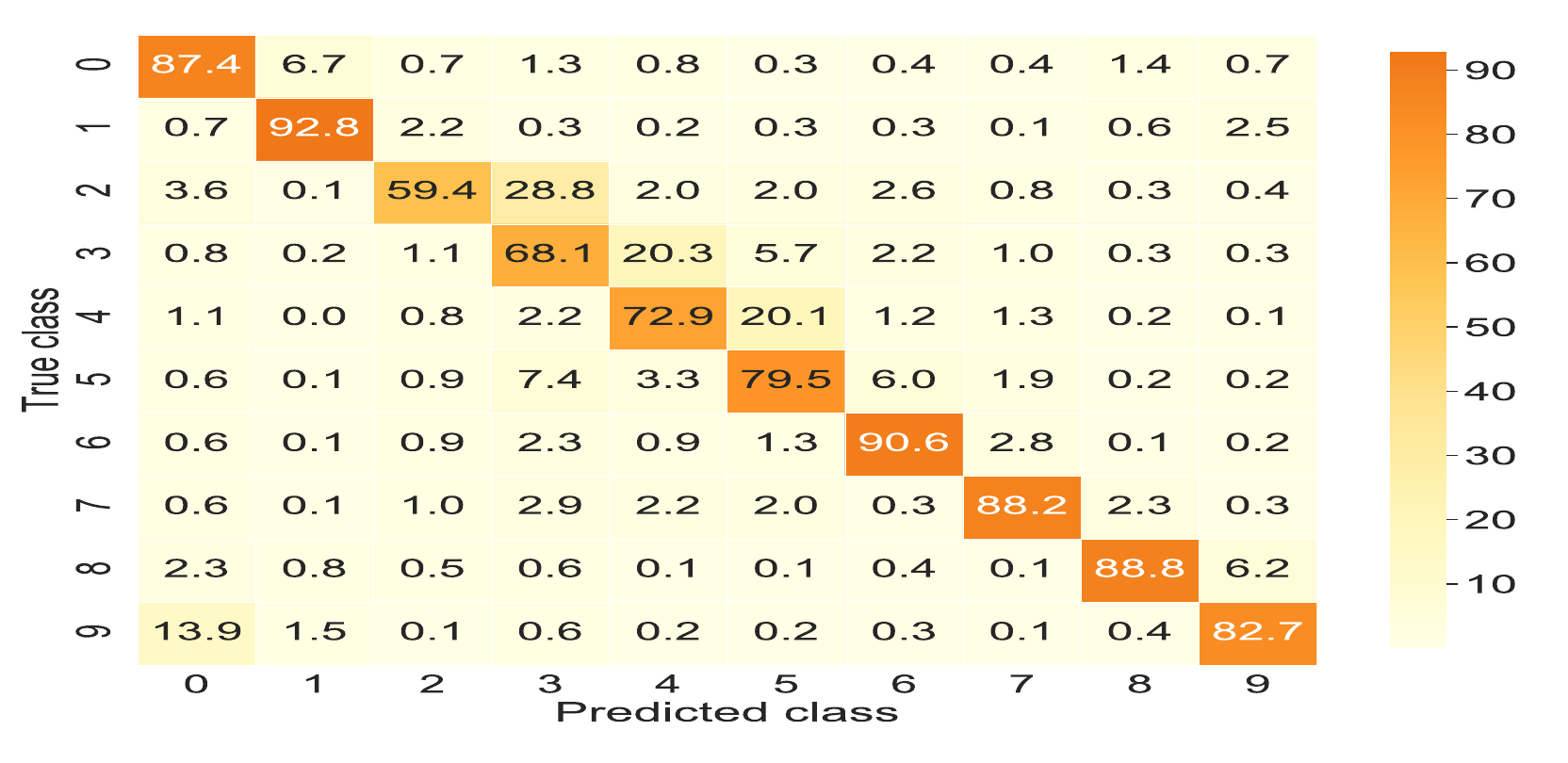}
		\caption{Confusion matrix.} \label{rr-1}
	\end{subfigure}
	\begin{subfigure}[b]{0.21\textwidth}
		\includegraphics[width=\textwidth]{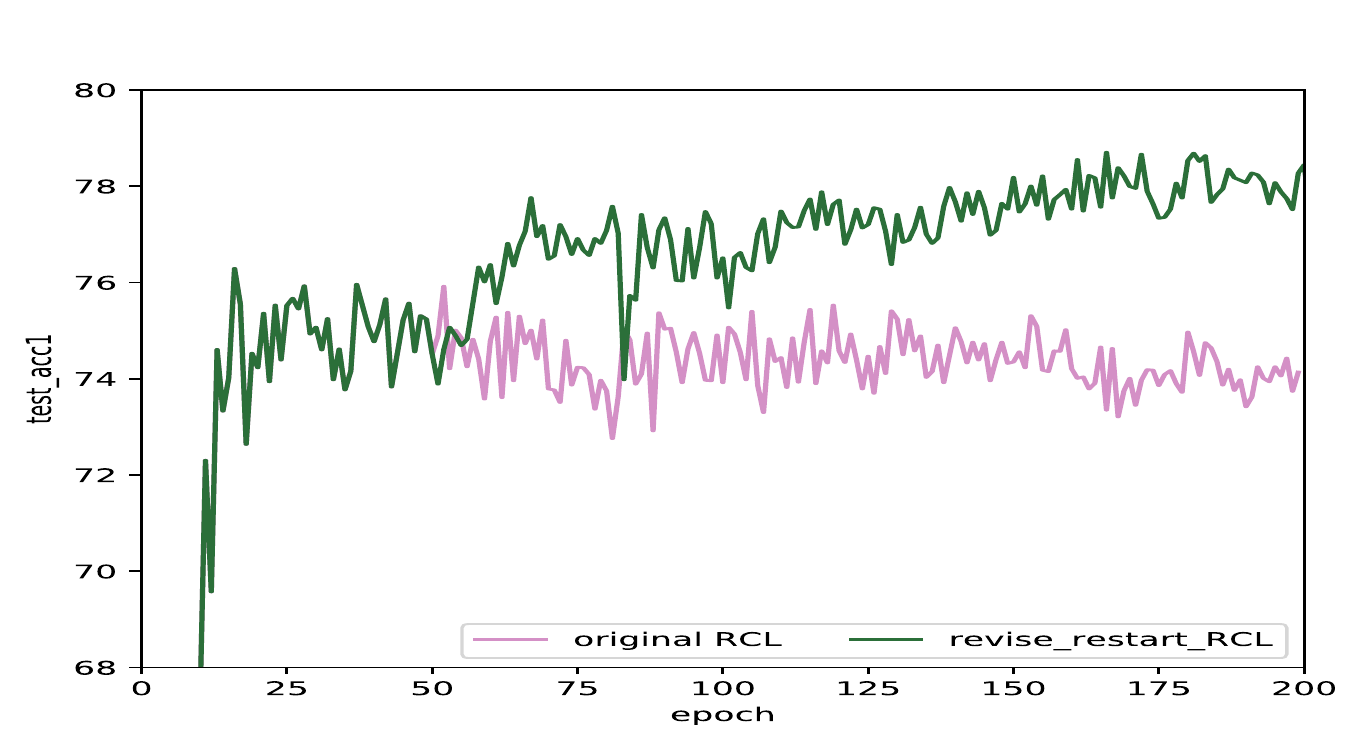}
		\caption{Test accuracy} \label{rr-2}
	\end{subfigure}
	\caption{Revise and restart RCL on CIFAR10 PF 45\% (K=3). Left: confusion matrix after revision. Numbers in cells denote percentages. Right: test accuracy w.r.t epochs. }\label{rr}
\end{figure}

\noindent\textbf{Reduce time complexity} While being effective, RCL requires more computational power and running time compared to methods using one or two networks. One way to reduce time complexity without deteriorating the performance is to utilize the unselected samples during training. Therefore, we propose a revise-and-restart strategy based on the current framework. When the training reaches certain epochs (usually plateau), we first revise the labels of the unselected samples based on the current prediction. Then we restart the training procedure, i.e., first introduce disagreement and then agreement. Fig.~\ref{rr} shows the result on CIFAR10 by using only 3 networks. After revising the labels of unselected samples at epoch 50, the label precision for each class is shown in Fig.~\ref{rr-1}, which is significantly higher compared to 55\%. Fig.~\ref{rr-2} shows the episode curve of revise and restart compared to the original 3 networks. The performance of revise-and-restart RCL using 3 networks reaches as high as that of 9 networks, which is a considerable reduction on the computational burden.

\begin{table*}[tb]
	\begin{tabular}{ccccccccccccccccccc}
		\hline \centering \small
		\multirow{2}{*}{\begin{tabular}[c]{@{}c@{}}Cell\\ Line\end{tabular}} & \multicolumn{2}{c}{Standard} & \multicolumn{2}{c}{SPL}   & \multicolumn{2}{c}{Decoupling} & \multicolumn{2}{c}{Co-teaching} & \multicolumn{2}{c}{SELF} & \multicolumn{2}{c}{LNEC} & \multicolumn{2}{c}{RCL} & p-val   & \multicolumn{2}{c}{Best} \\ \cline{2-18} 
		& ACC & F1 & ACC & F1 & ACC & F1 & ACC & F1 & ACC  & F1 & ACC & F1 & ACC & F1 & ACC & K & FR                               \\ \hline
		VCAP & 47.64 & 0.462 & 49.36 & 0.473 & 47.70 & 0.473 & 50.26 & 0.482 & 49.23 & 0.474 & 49.12 & 0.472 & \textbf{52.22} & \textbf{0.495} & \textbf{1e-3} & 4 & 0.3 \\
		MCF7 & 51.41 & 0.446 & 55.14 & 0.466 & 54.57 & 0.456 & 56.44 & 0.476 & 56.45 & 0.475 & 54.04 & 0.464 & \textbf{57.98} & \textbf{0.482} & \textbf{5e-4} & 4 & 0.2 \\
		PC3  & 47.90 & 0.474 & 47.92 & 0.473 & 45.32 & 0.467 & 48.22 & 0.477 & 48.42 & 0.478 & 48.18 & 0.475 & \textbf{49.04} & \textbf{0.484} & \textbf{8e-3} & 4 & 0.1 \\
		A549 & 50.73 & 0.403 & 53.52 & 0.417 & 53.38 & \textbf{0.422} & 52.15 & 0.414 & 53.67 & 0.421 & 53.63 & 0.417 & \textbf{54.00} & 0.421 & 0.25 & 3 & 0.1	\\
		A375 & 46.42 & 0.396 & 47.14 & 0.395 & \textbf{49.37} & \textbf{0.407} & 48.87 & 0.399 & 48.43 & 0.402 & 46.95 & 0.393 & 48.99 & 0.395 & - & 3 & 0.4  \\
		HT29 & 46.39 & 0.441 & 46.51 & 0.444 & 47.18 & 0.449 & 47.86 & 0.456 & 47.12 & 0.453 & 46.66 & 0.452 & \textbf{48.13} & \textbf{0.458} & \textbf{0.05} & 3 & 0.3 \\ \hline
	\end{tabular}
	\caption{Final performance for six cell lines (avg. 5 runs). K = number of networks, FR = forget rate. Significance $t$-tests (one-side) are conducted between RCL and the best baseline. A $p$-value less than 0.05 is considered as significant difference.}\label{real}
	\vspace{-2em}
\end{table*}

\subsection{Drug-induced Gene-Expression Change Prediction}

We apply RCL to a real-world problem in the bioinformatics domain, where the noise naturally exists and the noise rate is \textbf{unknown} (tuned as a hyper-parameter). 

Due to space limit, we briefly introduce the experimental settings. The task is to predict the responses of 11,000 drugs on 978 genes (down/up-regulate and no change, 3 classes) for 6 cell lines, whose results can be used for cancer drug discovery \cite{subramanian2017next, chen2017reversal}. The original responses are continuous values and each drug profile is repeatedly tested for different number of times. Some drugs profiles have consistent readings while others do not, indicating different label qualities. We obtain a small subset of high-quality profiles following standard statistical procedure \cite{subramanian2017next} to serve as test set and use the remaining data for training. The features and network architecture follow \cite{woo2019deepcop}. Table~\ref{real} shows the results in which RCL significantly outperforms baselines in most cases.

\section{Conclusion}
In this paper, we proposed a novel deep framework RCL for learning with noisy labels. Both synthetic and real experiments demonstrate the power of RCL. While being effective, RCL deserves further exploration and perfection in the future.


\section*{Acknowledgment}
This research is supported in part by NSF IIS-1749940 (JZ), ONR N00014-17-1-2265 (JZ), NIH 1R01GM134307 (JZ, BC), NIH K01ES028047 (BC).




%

\bibliographystyle{IEEEtran}
\bibliography{ref}

\end{document}